\documentclass{ecai} 

\usepackage{latexsym}
\usepackage{amssymb}
\usepackage{amsmath}
\usepackage{amsthm}
\usepackage{booktabs}
\usepackage{enumitem}
\usepackage{graphicx}
\usepackage{color}

\usepackage[dvipsnames]{xcolor}
\usepackage{makecell}
\usepackage{booktabs}
\usepackage{placeins}

\newcommand{\BibTeX}{B\kern-.05em{\sc i\kern-.025em b}\kern-.08em\TeX}

\begin{document}


\begin{frontmatter}


\paperid{336} 


\title{Data Selection: A General Principle for Building Small Interpretable Models}


 \author[A]{\fnms{Abhishek}~\snm{Ghose}}



\begin{abstract}
We present convincing empirical evidence for an effective and general strategy for building accurate small models. Such models are attractive for interpretability and also find use in resource-constrained environments. The strategy is to \emph{learn} the training distribution and sample accordingly from the provided training data. The distribution learning algorithm is not a contribution of this work; our contribution is a rigorous demonstration of the broad utility of this strategy in various practical settings. We apply it to the tasks of (1) building cluster explanation trees, (2) prototype-based classification, and (3) classification using Random Forests, and show that \emph{it improves the accuracy of decades-old weak traditional baselines to be competitive with specialized modern techniques}.

This strategy is also versatile wrt the notion of model size. In the first two tasks, model size is considered to be number of leaves in the tree and the number of prototypes respectively. In the final task involving Random Forests, the strategy is shown to be effective even when model size comprises of more than one factor: number of trees and their maximum depth.

Positive results using multiple datasets are presented that are shown to be statistically significant. 
\end{abstract}

\end{frontmatter}


\section{Introduction}

The application of Machine Learning to various domains often leads to specialized requirements. One such requirement is small model size, which is useful in the following scenarios:
\begin{enumerate}
    \item Interpretability: It is easier for humans to parse models when they are small. For example, a decision tree with a depth of 5 is likely easier to understand than one with a depth of 50. Multiple user studies have established model size as an important factor for interpretability \citep{feldman_boolean_coomplexity,Lage_Chen_He_Narayanan_Kim_Gershman_Doshi-Velez_2019,poursabzi-sangdeh2021manipulating}. This is also true for \emph{explanations} that are models, where \emph{post-hoc} interpretability is desired, e.g., local linear models with a small number of non-zero coefficients as in \emph{LIME} \citep{Ribeiro:2016:WIT:2939672.2939778} or  \emph{cluster explanation trees} with a small number of leaves \citep{imm_pmlr-v119-moshkovitz20a,ShallowTree_DBLP:journals/corr/abs-2112-14718}.  
    \item Resource-constrained devices: Small models are preferred when the compute environment is limited in various ways, such as memory and power \citep{tinyml_9166461,protoNN,machine_learning_at_edge}. Examples of such environments are micro-controllers, embedded devices and edge devices.
\end{enumerate}
Of course, in these cases, we also prefer that small models do not drastically sacrifice predictive power.
Typically this trade-off between size and accuracy is controlled in a manner that is specific to a model's formulation, e.g., $L1$ regularization for linear models, early stopping for Decision Trees. However, in this work we highlight a universal strategy: \emph{learn the training distribution}. This can often increase accuracy of small models, and thus addresses this trade-off in a model-agnostic manner. This was originally demonstrated in \citet{arxiv_learn_from_oracle,frontiers_density_tree}. Our work significantly extends the empirical evidence to show that such a strategy is indeed general and effective, and presents new evidence to show that it is also competitive.
\subsection{Motivation and Contributions}
\label{sec:contributions}
We empirically show that the strategy of learning the training distribution improves accuracy of small models in diverse setups. Previous work that proposed this technique \citep{arxiv_learn_from_oracle,frontiers_density_tree} demonstrate such improvements for a \emph{given} model; however, they leave an important question unanswered: are these improvements significant enough to be comparable to \emph{other} models \emph{tailored} to a task? Why might we use this strategy instead of using a contemporary specialized technique?  This is the practical gap this work fills, where our answer is affirmative.

We consider the following tasks, for which we set ourselves the goal of constructing easy-to-interpret models: 
\begin{enumerate}
    \item Building cluster explanation trees. 
    \item Prototype-based classification.
    \item Classification using Random Forests (RF).
\end{enumerate}

For evaluation on each of these tasks, we follow a common theme: (a) first, we show that a traditional technique is almost always not as good as newer and specialized techniques, and, (b) then we show that its performance may be radically improved by learning the training distribution. Collectively, these evaluations show that the strategy of learning the training distribution is both \emph{general} - may be applied to different tasks, models, notions of model sizes -  and \emph{effective} - results in competitive performance.  These rigorous evaluations in diverse practical setups is the primary contribution of this work.

\begin{table*}
  \caption{\textbf{Experiments at a glance}. The methods that are \textcolor{red}{highlighted in red} were augmented with training distribution learning in our experiments. The year in which a method was proposed is also mentioned. Notably, this strategy improves relatively old methods to be at par with newer specialized techniques for varied notions of model size. What ``model size'' means for a task is \textcolor{blue}{specified in blue}. The \emph{p}-value from a statistical test comparing the original and augmented models appears in the last column; smaller is better.}
  \label{tab:experiments_summary}
  \begin{tabular}{lllllll}
    \toprule
    Task & Methods & Metric & \thead{Significance\\Tests} & Datasets & \thead{\emph{p}-value of improvements\\(smaller is better)}\\
    \midrule
    \makecell{(1) Explainable Clustering\\ \textcolor{blue}{size\;=\;\# leaves in a tree}\\Section \ref{sec:expclust}} & \makecell{Iterative Mistake \\Minimization (\textcolor{CadetBlue}{2020}), \\ExShallow (\textcolor{CadetBlue}{2021}), \\ \textcolor{red}{CART} (\textcolor{red}{1984})} & \makecell{Cost Ratio \\ (ratio of F1\\ scores)} & \makecell{Friedman, \\ Wilcoxon} & \makecell{(1) avila, (2) Sensorless, \\ (3) covtype.binary, (4) covtype, \\ (5) mice-protein} & $p=1.4783 \times 10^{-6}$\\

    &&&&&&\\
    \makecell{(2) Prototype-based\\Classification\\\textcolor{blue}{size\; =\; \# prototypes}\\Section \ref{sec:proto}} & \makecell{ProtoNN (\textcolor{CadetBlue}{2017}),\\ Stochastic Neighbor\\Compression (\textcolor{CadetBlue}{2014}), \\Fast Condensed Nearest\\Neighbor Rule (\textcolor{CadetBlue}{2005})\\ \textcolor{red}{RBF Network} (\textcolor{red}{1988})} & F1-macro & \makecell{Friedman, \\ Wilcoxon} & \makecell{(1) adult, (2) covtype.binary, \\(3) senseit-sei, (4) senseit-aco\\(5) phishing} &  $p=1.699 \times 10^{-4}$\\

    &&&&&&\\
    \makecell{(3) Classification using \\Random Forests\\\textcolor{blue}{size\;=\;\{tree depth, \# trees\}}\\Section \ref{sec:rf}} & \makecell{Optimal Tree Ensemble (\textcolor{CadetBlue}{2020}), \\ subforest-by-prediction (\textcolor{CadetBlue}{2009}),\\ \textcolor{red}{Random Forest} (\textcolor{red}{2001})} & F1-macro & \makecell{Friedman, \\ Wilcoxon} & \makecell{(1) Sensorless, (2) heart, \\(3) covtype, (4) breast cancer,\\(5) ionosphere}&$p=1.44\times 10^{-11}$\\
    
    \bottomrule
  \end{tabular}
\end{table*}

Table \ref{tab:experiments_summary} concisely presents our setup and (one set of) observations. It lists various techniques compared on a task, including the traditional technique whose performance we seek to improve (highlighted in \textcolor{red}{red}), and the task-specific notion of model size (in \textcolor{blue}{blue}). For all evaluations multiple trials are  run (per dataset and model size combination; the datasets are also listed), and tests for statistical significance are conducted. The \emph{p}-value of the improvement in prediction accuracy (based on a \emph{Wilcoxon signed-rank test}) - of the traditional model using the provided training data vs using sampled data based on a learned distribution -  is shown in the last column  (smaller numbers are better). The general setup is further detailed in Section \ref{sec:common_expt_setup}.

\subsection{Organization}
This is conceptually a short paper, i.e., has a simple well-defined central thesis, and is predominantly an empirical paper, i.e., the thesis is validated using experiments. We begin with a discussion of prior work (Section \ref{sec:prev_work}), followed by a brief overview of the technique we use to learn the training distribution (Section \ref{sec:overview}). The latter provides relevant context for our experiments. Critical to this paper is our measurement methodology - this is described in Section \ref{sec:common_expt_setup}. The tasks mentioned in Table \ref{tab:experiments_summary} are respectively detailed in Sections \ref{sec:expclust}, \ref{sec:proto} and \ref{sec:rf}. We summarize various results and discuss future work in the Section \ref{sec:discussion_future_work}, which concludes the paper.

\section{Previous Work}
\label{sec:prev_work}
As mentioned, the only works we are aware of that discuss learning of the training distribution as a means to construct small models are \citet{arxiv_learn_from_oracle,frontiers_density_tree}. They propose techniques  that maybe seen as forms of ``adaptive sampling'': parameters for the training distribution are iteratively learned by adapting them to maximize  held-out accuracy.  In the rest of this paper, we will refer to these techniques as ``\textbf{Co}mpaction by  \textbf{A}daptive \textbf{S}ampling'' (\textbf{COAS}). 

\section{Overview of COAS}
\label{sec:overview}
COAS iteratively learns the parameters of a training distribution based on performance on a held-out subset of the data, given a training algorithm, desired model size $\eta$ and an accuracy metric of interest. It uses \emph{Bayesian Optimization (BO)} \citep{BO_overview} for its iterative learning. This specific optimizer choice allows for models with non-differentiable losses, e.g., decision trees. The output of one run of COAS is a \emph{sample} of the training data drawn according to the learned distribution. For our experiments, the following hyperparameters of COAS are relevant:
\begin{enumerate}
    \item Optimization budget, $T$: this is the number of iterations for which the optimizer runs. There is no other stopping criteria.
    \item The lower and upper bounds for the size of the sample to be returned. This sample size is denoted by $N_s$.
\end{enumerate}
Reasonable defaults exist for other hyperparameters \citep{arxiv_learn_from_oracle}. We use the reference library \emph{compactem} \citep{compactem2020} in our experiments.
A detailed review of COAS appears in Section \ref{appendix:coas} of the Appendix.

\section{Measurement}
\label{sec:common_expt_setup}

While each task-specific section contains a detailed discussion on the experiment setup, we discuss some common aspects here:
\begin{enumerate}
    \item To compare model families $\mathcal{F}_1, \mathcal{F}_2, \mathcal{F}_3$, each of which is, say, used to construct models for different sizes $\eta \in \{2,3\}$, for datasets $D_1, D_2, D_3$, we use the \emph{mean rank}, and support our conclusions with statistical tests such as the \emph{Friedman} \citep{friedman1} and \emph{Wilcoxon signed-rank} \citep{wilcoxon} tests\footnote{The \emph{Wilcoxon signed-rank} test was used here since it has been advocated by various studies  for measuring classification performance \citep{JMLR:v7:demsar06a,JMLR:v17:benavoli16a,japkowicz_shah_2011}.}. 
    
    Typically mean rank is used to compare model families based on their accuracies across datasets - which, ignoring model sizes,  may be visualized as a $3\times3$ table here, with rows representing datasets, and columns denoting model families - see Figure \ref{fig:eval_schematic}(a). An entry such as ``$D_2,\mathcal{F}_3$'' represents the accuracy (or some other metric) of a model from family $\mathcal{F}_3$ on dataset $D_2$. Models are ranked on a per-dataset basis, i.e., row-wise, and the average ranks  (computed per family, i.e., column-wise) are reported (lower is better). For statistical tests, the column values are directly used.

    However, we have an additional factor here - the model size. To avoid inventing a custom metric, we assimilate it in the previous scheme by using the combination of datasets \emph{and} model sizes as a row - see Figure \ref{fig:eval_schematic}(b). We think of such combinations as ``pseudo-dataset'' entries, i.e., now we have a $6 \times 3$ table, with rows for $D_1^2, D_1^3, D_2^2, D_2^3, D_3^2, D_3^3$, and same columns as before. The entry for ``$D_1^2, \mathcal{F}_3$'' indicates the accuracy of a model of size $2$ from family $\mathcal{F}_3$ on dataset $D_1$. 

    Effectively, now the comparisons automatically account for model size since we use pseudo-datasets instead of datasets.
    \textbf{Note} that no new datasets are being created - we are merely defining a convention to include model size in the familiar dataset-model cross-product table.
    
    \item For \emph{each} model family, model size and dataset combination (essentially a cell in this cross-product table), models are constructed multiple times (we refer to these as multiple ``trials''), and their scores are averaged. For tasks \#1 and \#2, five trials were used, whereas for task \#3, three trials were used. 
    
    
    \begin{figure}
    \centering
    \includegraphics[width=0.48\textwidth]{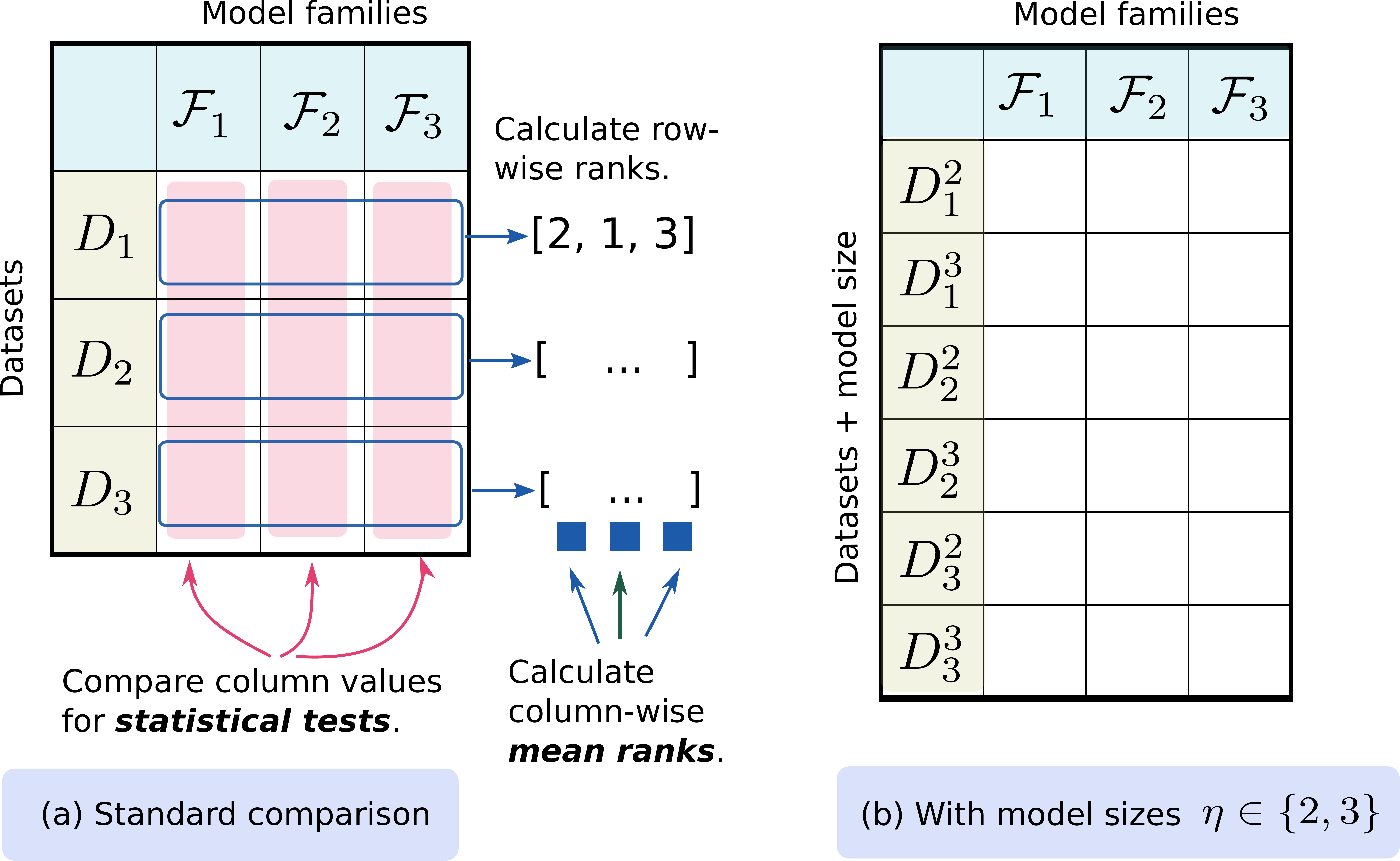} 
    \caption{(a) shows a standard measurement scheme with datasets in rows and model families in columns. Statistical tests are performed on the column values. Row-wise ranks are first computed for calculating the mean rank. (b) To account for model sizes, we allow rows to represent combinations of datasets and model sizes. See text for  details.  } 
    \label{fig:eval_schematic}
    \end{figure}

    \item Key results for tasks \#1,  \#2 and \#3 appear in Sections \ref{sec:expclust_obs}, \ref{sec:proto_obs} and \ref{sec:rf_obs} respectively. 
\end{enumerate}

\section{Explainable Clustering}
\label{sec:expclust}

The first task we investigate is the problem of \emph{Explainable Clustering}. Introduced by \cite{imm_pmlr-v119-moshkovitz20a}, the goal is to explain cluster allocations as discovered by  techniques such \emph{k-means} or \emph{k-medians}. This is achieved by constructing axis-aligned decision trees with leaves that either exactly correspond to clusters, e.g., \emph{Iterative Mistake Minimization (IMM)} \cite{imm_pmlr-v119-moshkovitz20a}, or are proper subsets, e.g., \emph{Expanding Explainable k-Means Clustering (ExKMC)} \cite{frost2020exkmc}. We consider the former case here, i.e., a tree must possess exactly $k$ leaves to explain $k$ clusters.

For a specific clustering $C$, let $C(x_i)$ denote the assigned cluster for an instance $x_i, i=1...N$,  where $C(x_i) \in \{1,2,...,k\}$, and the cluster centroids by $\mu_j, j=1,..., k$.  The cost of clustering  $J$ is then given by:
\begin{equation}
\label{eqn:kmeans}
    J = \frac{1}{N}\sum_{j=1}^k\sum_{\{x_i|C(x_i)=j\}} ||x_i - \mu_j ||_2^2
\end{equation}

In the case of an explanation trees with $k$ leaves, $\mu_j$ are centroids of leaves. Cluster explanation techniques attempt to minimize this cost. 

The price of explainability maybe measured as the \emph{cost ratio}\footnote{This is referred to as the \emph{cost ratio} in \citet{frost2020exkmc}, \emph{price of explainability} in \citet{imm_pmlr-v119-moshkovitz20a} and \emph{competitive ratio} in \citet{10.1145/3519935.3520056}.}:
\begin{equation}
    \label{eqn:cost_ratio}
    \text{cost ratio} = \frac{J_{Ex}}{J_{KM}}
\end{equation}

Here $J_{Ex}$ is the cost achieved by an explanation tree, and $J_{KM}$ is the cost obtained by a standard k-means algorithm. 
It assumes values in the range $[1, \infty]$, where the lowest cost is obtained when using k-means, i.e., $J_{Ex}$ and $J_{KM}$ are the same.   

One may also indirectly minimize the cost in the following manner: use k-means to produce a clustering, use the cluster allocations of instances as their labels, and then learn a standard decision tree for classification, e.g., CART. This approach has been shown to be often outperformed by tree construction algorithms that directly minimize the cost in Equation \ref{eqn:kmeans}, e.g., IMM.

\begin{figure*}
\centering
\includegraphics[width=0.9\textwidth]{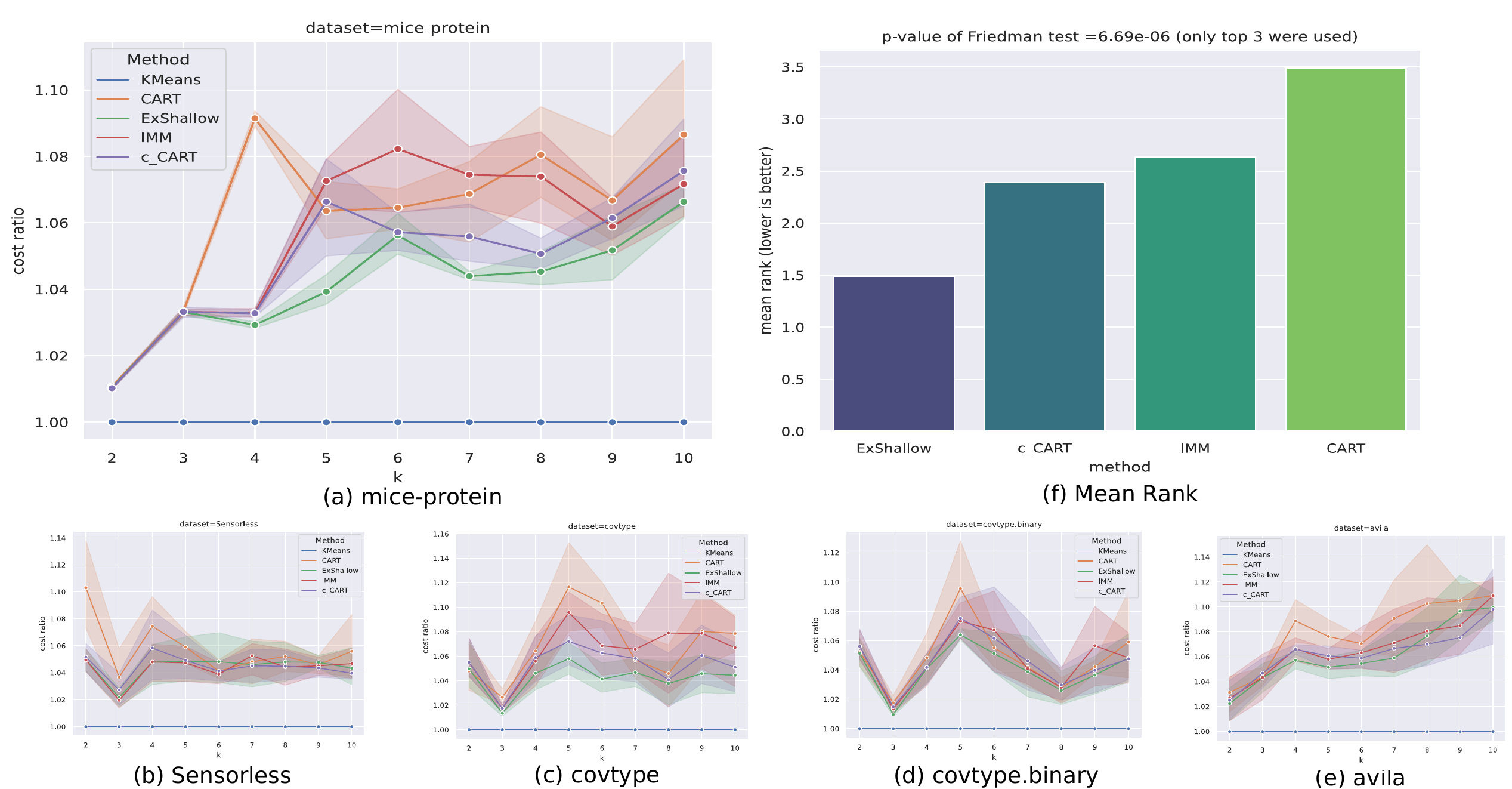} 
\caption{Comparisons over explainable clustering algorithms are shown. (a) shows the comparison for a specific dataset \emph{mice-protein}. (b), (c), (d) and (e) show comparisons over other datasets - miniaturized to fit the page. (f)  shows mean ranks of these techniques over five datasets across model sizes; the Friedman test is conducted over the \textbf{top three} techniques only, with  $p = 6.688 \times {10}^{-6}$. }. 
\label{fig:expclust}
\end{figure*}

\subsection{Algorithms and Hyperparameters}
\label{sec:expclust_algo_hyp}
The algorithms we compare and their hyperparameter settings are as follows:
\begin{enumerate}
    \item \textbf{Iterative Mistake Minimization (IMM)} \cite{imm_pmlr-v119-moshkovitz20a}: This generates a decision tree via greedy partitioning using a criterion that minimizes number of mistakes at each split (the number of points separated from their corresponding reference cluster center). There are no parameters to tune. We used the implementation available here: \url{https://github.com/navefr/ExKMC}, which internally uses the reference implementation for IMM.
    \item \textbf{ExShallow} \cite{ShallowTree_DBLP:journals/corr/abs-2112-14718}: Here, the decision tree construction explicitly accounts for minimizing explanation complexity while targeting a low cost ratio. The trade-off between clustering cost and explanation size is controlled via a parameter $\lambda$. This is set as $\lambda=0.03$ in our experiments; this value is used in the original paper for various experiments. We used the reference implementation available here: \url{https://github.com/lmurtinho/ShallowTree}.
    \item 
\textbf{CART} w/wo \textbf{COAS}: We use CART \cite{cart93} as the traditional model to compare, and maximize the classification accuracy for predicting clusters, as measured by the F1-macro score. The implementation in \emph{scikit} \cite{scikit-learn} is used. During training, we set the following parameters: (a) the maximum number of leaves (this represents \emph{model size} $\eta$ here) is set to the number of clusters $k$, and (b) the parameter \emph{class\_weight} is set to \emph{``balanced''} for robustness to disparate cluster sizes. Results for CART are denoted with label \textbf{CART}. We then apply COAS to CART; these results are denoted as \textbf{c\_CART}. We set $T=2000$, and use default settings for other parameters, e.g., $N_s \in [400, |X_{train}|]$. Since we are explaining clusters (and not predicting on unseen data), the training, validation and test sets are identical. 
\end{enumerate}

\subsection{Experiment Setup}
The comparison is performed over five datasets (limited to $1000$ instances), and for each dataset,  $k=2,3,...,10$ clusters are produced. Results for the cost ratio (Equation \ref{eqn:cost_ratio}) are reported over \emph{five} trials. Evaluations are performed over the following publicly available datasets: \emph{avila}, \emph{covtype}, \emph{covtype.binary}, \emph{Sensorless} \cite{libsvm} and \emph{mice-protein} \cite{Dua:2019}. We \emph{specifically picked} these datasets since CART is known to perform poorly on them \cite{frost2020exkmc,ShallowTree_DBLP:journals/corr/abs-2112-14718}, and thus these provide a good opportunity to showcase the power of this strategy.

\subsection{Observations}
\label{sec:expclust_obs}
Figure \ref{fig:expclust} presents our results. Figure \ref{fig:expclust}(a) shows the plot for the \emph{mice-protein} dataset: the $95\%$ confidence interval, in addition to cost ratio, is shown\footnote{It might come as a surprise that the cost ratio increases with increasing $k$, but this seems to be a transient phenomenon; at even higher values of $k$ we do observe that cost ratios collectively decrease. See Section \ref{appendix:derease_cost_ratio}, Appendix.}. Plots for other datasets are shown miniaturized - (b), (c), (d), (e) in the interest of space. The cost for k-means is shown for reference a blue horizontal line at $y=1$. Figure \ref{fig:expclust}(f) shows the \emph{mean ranks} of the various techniques (lower is better) across datasets and number of clusters (as described in Section \ref{sec:common_expt_setup}, trials scores are averaged),  and its title shows the \emph{p-value}$=6.688 \times 10^{-6}$ of a \emph{Friedman test} conducted over \emph{the top three techniques}: we restrict the test to top candidates since otherwise it would be very easy to obtain a low score favorable to us, due to the high cost ratios for CART. The low score indicates with high confidence that ExShallow, IMM and c\_CART do not produce the same outcomes.

From the plot of mean ranks in Figure \ref{fig:expclust}(f), we observe that although CART performs quite poorly, the application of COAS drastically improves its performance, to the extent that it competes favorably with techniques like IMM and ExShallow; its mean rank places it between them. This is especially surprising given that it doesn't explicitly minimize the cost in Equation \ref{eqn:kmeans}. We also note the following \emph{p-values} from \emph{Wilcoxon signed-rank} tests:
\begin{itemize}
    \item CART vs c\_CART: $p=1.4783 \times 10^{-6}$. The low value indicates that using COAS indeed significantly changes the accuracy of CART.
    \item IMM vs c\_CART: $p=0.0155$. The relatively high value indicates that the performance of c\_CART is competitive with IMM.
\end{itemize}

Here, both the Friedman and Wilcoxon tests are performed for combinations of datasets and $k$ - a ``pseudo-dataset'', as discussed in Section \ref{sec:common_expt_setup}.

\begin{figure*}[h]
\centering
\includegraphics[width=0.9\textwidth]{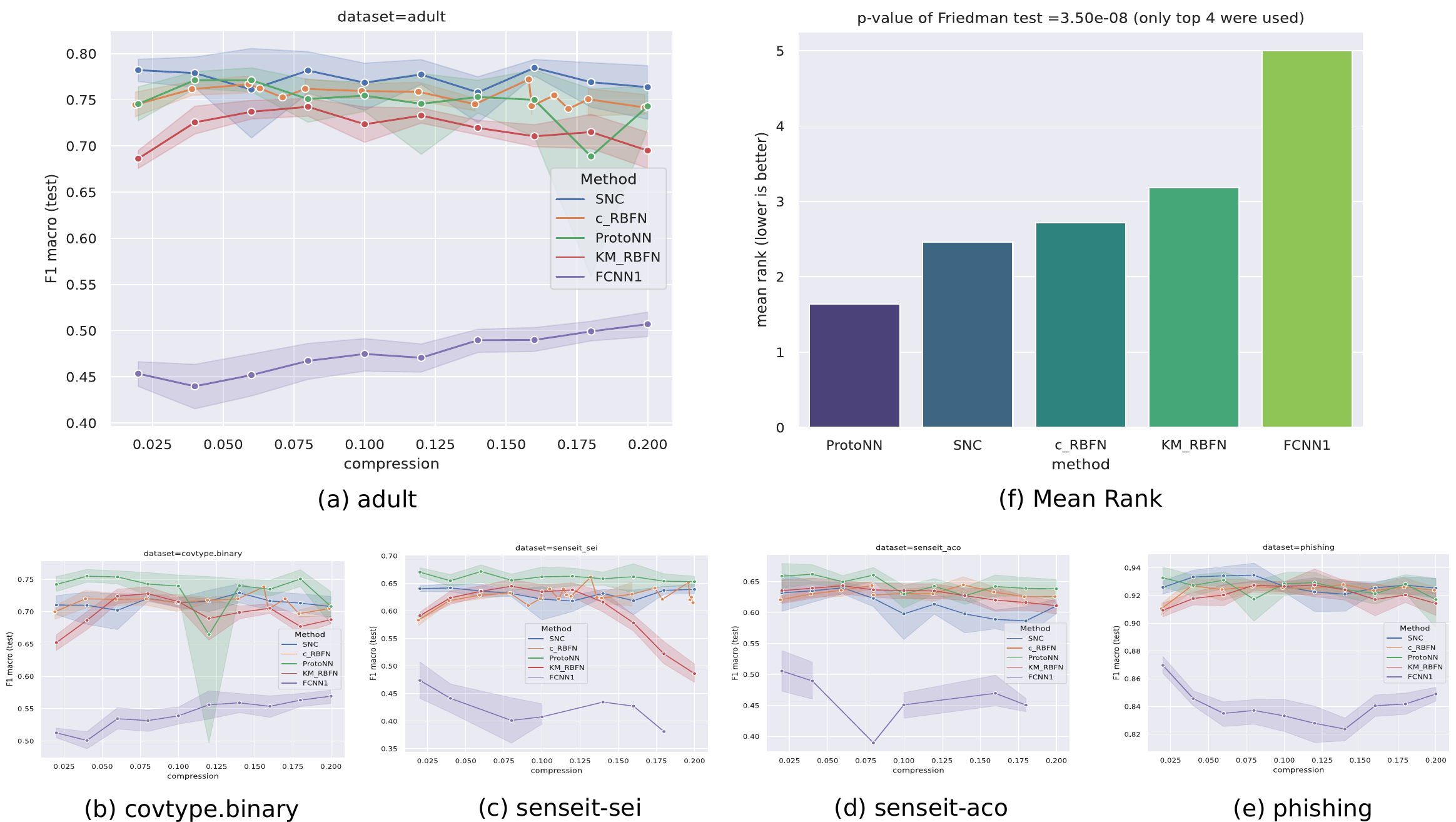} 
\caption{Various prototype-based classifiers are compared. (a) shows comparison for the dataset \emph{adult}. Number of prototypes are shown as percentage of the training data on the \emph{x-axis}, and is referred to as ``compression''. (b), (c), (d) and (e) shows plots for other datasets - these are miniaturized to fit the page. (f) shows the mean ranks of techniques based on five datasets; the Friedman test is conducted over the \textbf{top four} techniques only, with  $p=3.5025\times 10^{-8}$. }.
\label{fig:proto}
\end{figure*}

\section{Protoype-based Classification}
\label{sec:proto}
Next, we consider prototype-based classification. At training time, such techniques identify ``prototypes'' (actual training instances or generated instances), that maybe used to classify a test instance based on their similarity to them. A popular technique in this family is the \emph{k-Nearest Neighbor (kNN)}. These are simple to interpret, and if a small but effective set of protoypes maybe identified, they can be convenient to deploy on edge devices \cite{protoNN,9001024}.  Prototypes also serve as minimal ``look-alike'' examples for explaining models \citep{dl_proto,image_proto}. Research in this area has focused on minimizing the number of prototypes that need to be retained while minimally trading off accuracy.

We define some notation first. The number of prototypes we want is an input to our experiments, and is denoted by $N_p$. We will also use $K_{\gamma}(x_i, x_j) = e^{-\gamma||x_i-x_j ||_2^2} $ to denote the \emph{Radial Basis Function (RBF) kernel}, parameterized by the kernel bandwidth $\gamma$.  


\subsection{Algorithms and Hyperparameters}
\label{sec:proto_algo_hyp}
These are the algorithms we compare:
\begin{enumerate}
    
    \item \textbf{ProtoNN} \cite{protoNN}: 
    This technique uses a RBF kernel to aggregate influence of prototypes. Synthetic prototypes are learned and additionally a ``score'' is learned for each of them that designates their contribution towards \emph{each} class. The prediction function sums the influence of neighbors using the RBF kernel, weighing contribution towards each class using the learned score values; the class with the highest total score is predicted. The method also allows for reducing dimensionality, but we don't use this aspect\footnote{The implementation provides no way to switch off learning a projection, so we set the dimensionality of the projection to be equal to the original number of dimensions. This setting might however learn a transformation of the data to space within the same number of dimensions, e.g., translation, rotation.}. The various parameters are learned via gradient based optimization. 
    
    
    We use the \emph{EdgeML} library \citep{edgeml04}, which contains the reference implementation for ProtoNN. For optimization, the implementation uses the version of \emph{ADAM} \cite{DBLP:journals/corr/KingmaB14} implemented in \emph{TensorFlow} \cite{tensorflow2015-whitepaper}; we set $num\_epochs=200$, $learning\_rate=0.05$, while using the defaults for other parameters. The $num\_epochs$ and $learning\_rate$ values are picked based on a limited search among values $\{100, 200, 300\}$ and $\{0.01, 0.05\}$ respectively. The search space explored for $\gamma$ is $[0.001, 0.01, 0.1, 1, 10]$. Defaults are used for the other ProtoNN hyperparameters.
    

    \item \textbf{Stochastic Neighbor Compression (SNC)} \cite{snc}: This also uses a RBF kernel to aggregate influence of prototypes, but unlike ProtoNN, the prediction is performed via the \emph{1-NN rule}, i.e., prediction uses only the nearest prototype. The technique bootstraps with randomly sampled $N_p$ prototypes (and corresponding labels) from the training data, and then modifies their coordinates for greater accuracy using gradient based optimization; the labels of the prototypes stay unchanged in this process. This is another difference compared to ProtoNN, where in the latter, each prototype contributes to all labels to varying extents.  The technique maybe extended to reduce the dimensionality of the data (and prototypes); we don't use this aspect. 
    
    We were unable to locate the reference implementation mentioned in the paper, so we implemented our own version, with the help of the \emph{JAXopt} library \cite{jaxopt_implicit_diff}. For optimization, gradient descent with \emph{backtracking line search} is used. A total of $100$ iterations for the gradient search is used (based on a limited search among these values: $\{100, 200, 300\}$), and each backtracking search is allowed up to $50$ iterations. A grid search over the following values of $\gamma$ is performed: $[0.001, 0.01, 0.1, 1, 10]$.

    
    
    \item \textbf{Fast Condensed Nearest Neighbor Rule} \cite{fcnn}: Learns a ``consistent subset'' for the training data: a subset such that for any point in the training set (say with label $l$), the closest point in this subset also has a label $l$. Of the multiple variations of this technique proposed in \cite{fcnn}, we use \textbf{FCNN1}, which uses the \emph{1-NN} rule for prediction. There are no parameters to tune. We used our own implementation.
    
    A challenge in benchmarking this technique is it \emph{does not} accept $N_p$ as a parameter; instead it iteratively produces expanding subsets of prototypes until a stopping criteria is met, e.g., if prototype subsets $V_{i}$ and $V_{i+1}$ are produced at iterations $i$ and $i+1$ respectively, then they satisfy the relationship $V_{i}  \subset V_{i+1}$. For comparison, we consider the performance at iteration $i$ to be the result of $N_p$ prototypes where $N_p$ is defined to be $|V_i|$, i.e., instead of setting $N_p$, we use the value the algorithm produces at each iteration.
    
    \item \textbf{RBFN} w/wo \textbf{COAS}: 
    For the traditional model, we use  \emph{Radial Basis Function Networks (RBFN)} \cite{rbfn}.
    For a binary classification problem with classes $\{-1, 1\}$, given prototypes $x_i, i=1,2,...,p$, the label of a test instance $x$ is predicted as $ sgn(\sum_{i}^p w_i K_{\gamma}(x, x_i))$ (a score of $0$ is set to a label of $1$). Weights $w_i$ are learned using linear regression. A one-vs-rest setup is used for multiclass problems. For our baseline, we use cluster centres of a \emph{k-means} clustering as our prototypes, where $k$ is set to $N_p$. These results are denoted using the term \textbf{KM\_RBFN}. In the COAS version, denoted by \textbf{c\_RBFN}, the $N_p$ prototypes are sampled from the training data. $N_p$ represents \emph{model size} $\eta$ here.
    
    Note that the standard RBFN, and therefore the variants used here KM\_RBFN and c\_RBFN, don't provide a way to reduce dimensionality; this is the reason why this aspect of ProtoNN and SNC wasn't used (for fair comparison).
    
    For COAS, we set $T=1000$ and $N_s$ was set to $[N_p-1, N_p]$  to get the desired number of prototypes\footnote{The implementation \cite{compactem2020} doesn't allow for identical lower and upper bounds, hence the lower bound here is $N_p-1$.}.

    
\end{enumerate}

Although all the above techniques use prototypes for classification, it is interesting to note variations in their design: ProtoNN, SNC, KM\_RBFN use synthetic prototypes, i.e., they are not part of the training data, while c\_RBFN and FCNN1 select $N_p$ instances from the training data. The prediction logic also differs: ProtoNN, KM\_RBFN, c\_RBFN derive a label from some function of the influence by all prototypes, while SNC and FCNN1 use the 1-NN rule.


\subsection{Experiment Setup}
As before, we evaluate these techniques over five standard datasets: \emph{adult, covtype.binary, senseit-sei, senseit-aco, phishing} \cite{libsvm}. $1000$ training points are used, with  $N_p \in \{20, 40, 60, 80, 100, 140, 160, 180, 200\}$. Results are reported over five trials. The score reported is the F1-macro score.


\subsection{Observations}
\label{sec:proto_obs}
Results are shown in Figure \ref{fig:proto}. (a) shows the plot for the \emph{adult} dataset. The number of prototypes  are shown on the \emph{x-axis} as \emph{percentages} of the training data. Plots for other datasets are shown in (b), (c), (d) and (e); these have been miniaturized to fit the page. Figure \ref{fig:proto}(f) shows the mean rank (lower is better) across datasets and number of prototypes (as described in Section \ref{sec:common_expt_setup}, trials are aggregated over). The p-value of the Friedman test is reported, $p=3.5025\times 10^{-8}$. Here too, we do not consider the worst performing candidate,  FCNN1 - so as to not bias the Friedman test in our favor.

We observe in Figure \ref{fig:proto}(f) that while both ProtoNN and SNC outperform c\_RBFN, the performance of SNC and c\_RBFN are close. We also observe that FCNN1 performs poorly; this matches the observations in \citet{snc}.

We also consider the following \emph{p-values} from \emph{Wilcoxon signed-rank} tests:
\begin{enumerate}
    \item KM\_RBFN vs c\_RBFN: $p=1.699 \times 10^{-4}$. The low value indicates COAS significantly improves upon the baseline KM\_RBFN.
    \item  SNC vs c\_RBFN: $p=0.1260$. The relatively high value here indicates that c\_RBFN is competitive with SNC; in fact, at a confidence threshold of $0.1$, their outcomes would not be interpreted as significantly different.
\end{enumerate}
As discussed in Section \ref{sec:common_expt_setup}, these statistical tests are conducted over a combination of dataset and model size.

\section{Random Forest}
\label{sec:rf}

\begin{figure*}[ht]
  \centering
  \includegraphics[width=0.9\textwidth]{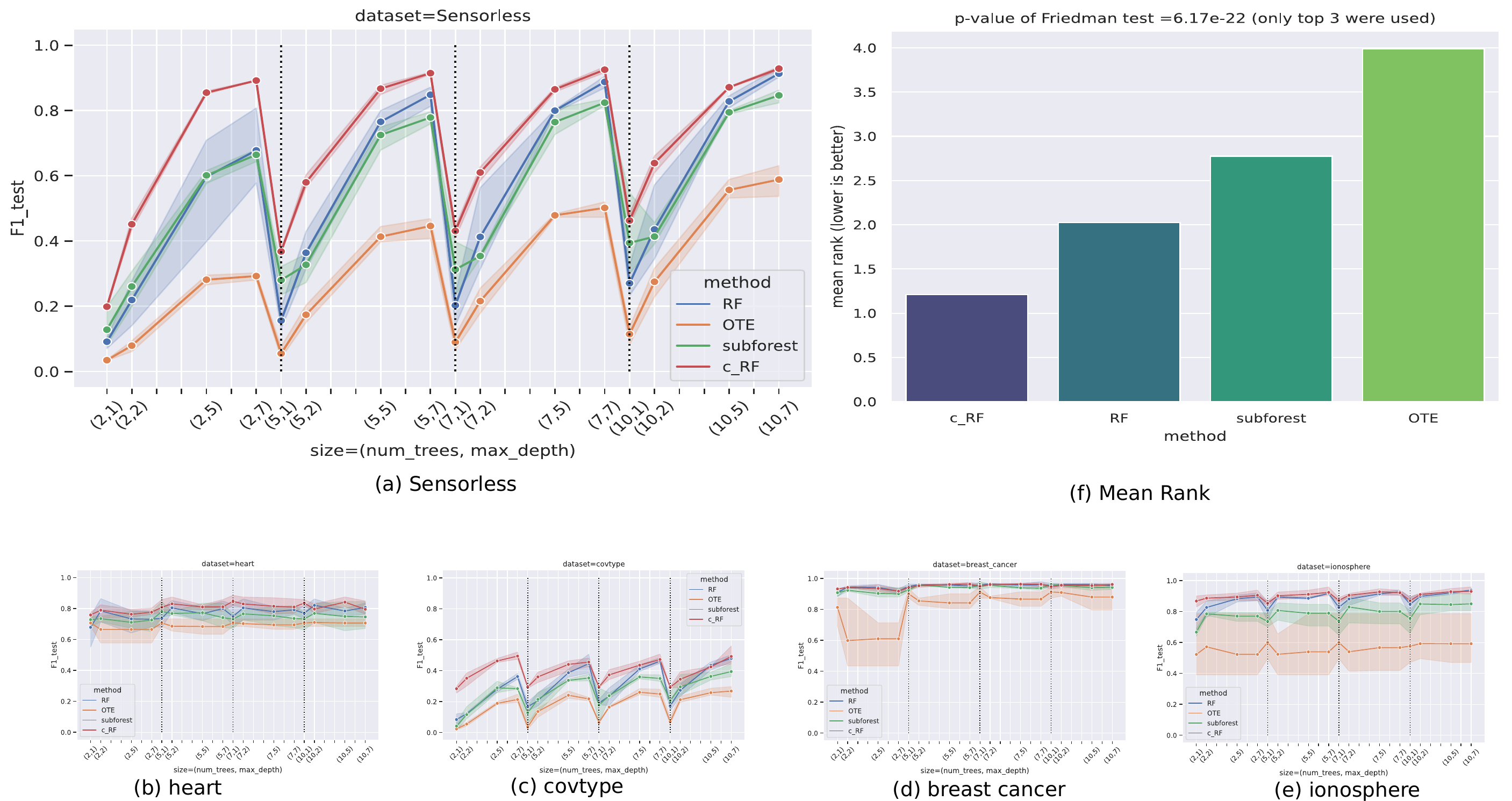}
  \caption{(a) shows results for the dataset=\emph{covtype}. (b), (c), (d) and (e) show plots for other datasets - miniaturized to fit the page. (f) provides mean ranks, but since there are only two models being compared, the Friedman test cannot be performed.}
\label{fig:rf}
\end{figure*}

In the previous sections, we considered the case of scalar model sizes: number of leaves in the case of explanation trees for clustering (Section \ref{sec:expclust}) and number of prototypes in the case of prototype-based classification (Section \ref{sec:proto}). Here, we assess the effectiveness of the technique when the model size is composed of multiple factors. 

We look at the case of learning \emph{Random Forests (RF)} where we specify model size using \emph{both} the number of trees and the maximum depth per tree.

\subsection{Algorithms and Hyperparameters}
\label{sec:algo_hyper_rf}
The algorithms compared are:
\begin{enumerate}
    \item \textbf{Optimal Tree Ensembles (OTE)} \citep{OTE_Khan2020}: This is a technique to \emph{prune} a RF to produce a lower number of trees. The pruning occurs in two phases: (a) first, the top $M$ trees are retained based on prediction accuracy on \emph{out-of-bag} examples, and (b) further pruning is performed based on a tree's contribution to the overall \emph{Brier score} on a validation set. In our experiments, $M$ was set to $20\%$ of the initial number of trees in the forest. 
    
    Although a reference \texttt{R} package exists \citep{OTE_R}, it doesn't allow to set the max. depth of trees, which is relevant here. Hence we use our implementation. 

    \item \textbf{Finding a subforest} \citep{subforest}: This paper proposes multiple techniques to prune the number of trees in an RF. Among them it shows that pruning based on incremental predictive power of trees performs best. Hence, this is the technique we use here, based on our own implementation.
    
    \item \textbf{RF} w/wo \textbf{COAS}: We train standard RFs \citep{Breiman2001_rf} without and with COAS, denoted as \textbf{RF} and \textbf{c\_RF} respectively. For COAS, we set number of iterations as $T=3000$ and $N_s \in [30, |X_{train}|]$ (the lower bound was obtained by limited search). The implementation in \emph{scikit} is used, and for standard RF models, i.e., without COAS, the parameter \emph{class\_weight} is set to \emph{``balanced\_subsample''} to make them robust to class imbalance.
\end{enumerate}

\subsection{Experiment Setup}
We use the following five standard datasets for this experiment:  \emph{heart, Sensorless, covtype, ionosphere, breast cancer} \cite{libsvm}. For each dataset, all size combinations from the following sets are tested (three trials per combination): $num\_trees \in \{2,5,7,10\}$ and \\ $max\_depth \in \{1,2,5,7\}$. The total dataset size used had $3250$ instances, with $70:30$ split for train to test. For both OTE and the subforest-finding technique, the RFs were originally constructed with $100$ trees, which were then pruned to the desired number. We report the F1-macro score in our experiments.

\subsection{Observations}
\label{sec:rf_obs}
Figure \ref{fig:rf}(a) shows the results for dataset \emph{Sensorless}; the \emph{x-axis} shows values of the tuple $(num\_trees, max\_depth)$ sorted by the first and then the second index. Note the ``sawtooth'' pattern. For a given value of $num\_trees$, increase in $max\_depth$ leads to increasing accuracy. However, as we move to the next value for $num\_trees$, we start with $max\_depth=1$ again, which leads to a drop in accuracy. This is an artefact of the ordering of the \emph{x-axis}, and is expected behavior.  Plots for other datasets are shown in (b), (c), (d) , (e) - these are miniaturized in the interest of space. Figure \ref{fig:rf}(f) shows mean ranks across datasets, number of trees and maximum tree depths (across trials). 

OTE results are discounted for the Friedman test on account of its relatively poor performance. This gives us a p-value of  $p=6.17\times 10^{-22}$, denoting sufficiently different outcomes.

It is easy to see that c\_RF is the best performing technique here. The \emph{p-value} for \emph{Wilcoxon signed-rank} test wrt the standard RF is $p=1.44\times 10^{-11}$, demonstrating significant improvement. 

In the interest of fairness, \textbf{please note} that both OTE and subforest-finding, in their original papers \emph{do not} restrict the max. depth of trees, and as such, their use here should be seen as a modification, and not indicative of their performance on the original intended use-cases.

\section{Discussion and Future Work}
\label{sec:discussion_future_work}
We note that a different number of iterations, $T$, were used for the tasks - see Sections \ref{sec:expclust_algo_hyp}, \ref{sec:proto_algo_hyp} and \ref{sec:algo_hyper_rf}. This is because there is no widely-accepted stopping criteria for BO (although it has been studied \citep{Makarova2021,pmlr-v206-ishibashi23a}), and the task-specific values were arrived at by limited search.  


The experiments here clearly showcase both the versatility and effectiveness of the  strategy of learning the training distribution: it may be applied to different models and notions of model sizes, and \emph{importantly}, it can produce results that are competitive with specialized techniques.

While not directly related, it is equally important for us to call out what we are \emph{not} claiming. We don't propose that COAS replace the techniques it was compared to without further study, as the latter may offer other task specific benefits, e.g., ExShallow targets other  explanation-quality metrics aside from the cost ratio. COAS might be able to address these dimensions as well, but it is beyond the scope of this study. Another reason for caution is that the improvements of COAS diminish as model sizes increase \citep{arxiv_learn_from_oracle} ; hence, its utility to a task depends on what model size range is acceptable.

Instead, this work is a call to action to explore the strategy of learning the training distribution to build small accurate models - which we believe is relatively unknown today.

In terms of future work, some directions are: (a) learn weights for training instances (as opposed to a distribution) using \emph{bilevel optimization} \cite{hoag,bilevel_hyperopt} for the case of a differentiable training loss function, (b) COAS itself may be improved with the use of a different black-box optimizer, and (c) develop a theoretical framework that explains this strategy.  





\bibliography{refs}

\clearpage

\appendix
\section{Appendix}
\subsection{Review of COAS}
\label{appendix:coas}
We briefly discuss COAS in this section. Specifically, we discuss the technique from \citet{arxiv_learn_from_oracle}\footnote{The techniques in the papers vary in terms of how they make the process of adaptive sampling tractable. In \citet{frontiers_density_tree}, a decision tree is used capture neighborhood information, which is then utilized for sampling. \citet{arxiv_learn_from_oracle} uses the \emph{prediction uncertainty} from an \emph{oracle} model to aid sampling. The latter was shown to be more accurate in the respective paper, and this is why we use it here.}. We simplify various aspects for brevity - for details please refer to the original paper.

We denote the \emph{sampling process} used in COAS by $S((X, Y), \Psi, N_s, p_o)$. Here, $(X, Y)$ is the data that is to be sampled from (\emph{with replacement}) and $\Psi, N_s, p_o$ are parameters used for sampling. These are defined as follows:
\begin{itemize}
    \item $\Psi$: This is a \emph{probability density function (pdf)} defined over the data\footnote{The \emph{pdf} is applies to the data \emph{indirectly}: it models the density of the prediction uncertainty scores of instances in $(X, Y)$, as provided by the oracle model. We use this simplification for brevity.} $(X, Y)$ and denotes the sampling probability of instances. 
    \item $N_s$: Sample size.
    \item $p_o$: The fraction of samples that are to be sampled uniformly randomly from $(X, Y)$. The remaining $(1-p_o) N_s$ samples are chosen with replacement from $(X, Y)$ based on the probabilities $p((X, Y); \Psi)$.
\end{itemize}

$p_o$ serves as a ``shortcut'' for the model to combine data that was provided with data from the learned distribution. Since the train and test data splits provided are assumed to come from the same distribution, $p_o$ also acts as tool for analyzing the ``mix'' of data preferred at various model sizes. For example, an interesting result shown \citep{arxiv_learn_from_oracle} is that as model size increases, $p_o \rightarrow 1$, i.e., the best training distribution at large sizes is the test distribution, which is the commonly known case.

To state the optimize problem COAS solves, we introduce additional notation:
\begin{enumerate}
    \item Let $(X_{train}, Y_{train})$ and $(X_{val}, Y_{val})$ represent training and validation datasets respectively.
    \item Let $train_{\mathcal{F}}(\eta, (X, Y))$ be a training algorithm that returns a model of size $\eta$ from model family $\mathcal{F}$ when supplied with training data $(X, Y)$.
    \item Let $acc(f, (X, Y))$ denote the accuracy of model $f$ on dataset $(X, Y)$.
\end{enumerate}
Then, COAS performs the following optimization for user-specified $\eta$ over $T$ iterations:
\begin{align}
    & \max_{\Psi, N_s, p_o} \; acc(f, (X_{val}, Y_{val})) \\
    \text{where, } & f = train_{\mathcal{F}}(\eta, (X_{sample}, Y_{sample}) \\
    \text{and } &(X_{sample}, Y_{sample}) = S((X_{train}, y_{train}), \Psi, N_s, p_o)
\end{align}
Essentially, COAS identifies parameters $\Psi, N_s, p_o$ such that a model trained on data sampled using them, maximizes validation accuracy. Note that since $S()$ samples with replacement, it is possible to have an optimal $N_s$ that's larger thann the size of the training data. A \emph{Bayesian Optimizer} is used for optimization, and a \emph{budget} of $T$ iterations is provided. There is no other stopping criteria, i.e., the optimizer is run all the way through $T$ iterations. 



\subsection{Cost Ratio Revisited}
\label{appendix:derease_cost_ratio}
\begin{figure}
    \centering
    \includegraphics[width=0.48\textwidth]{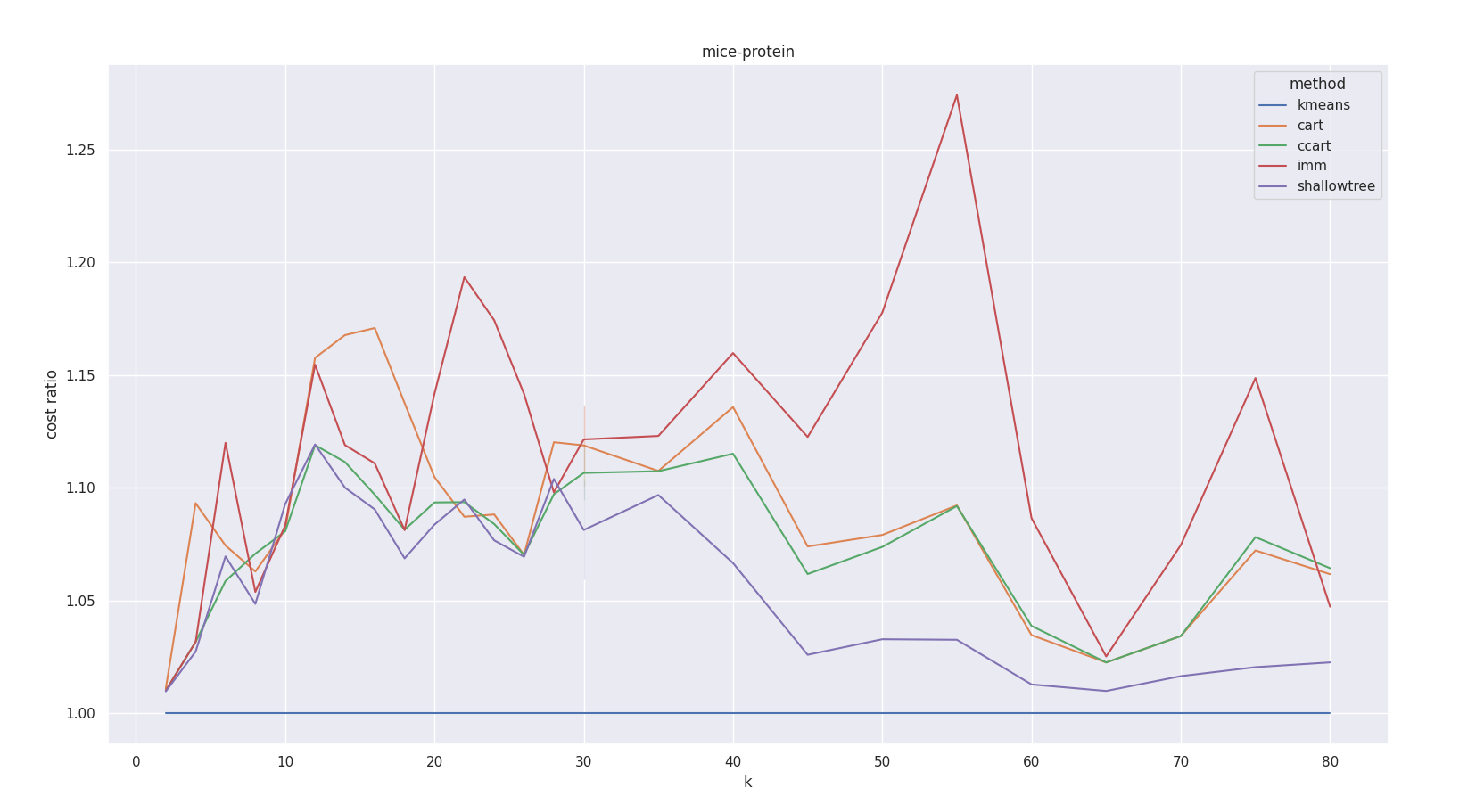} 
    \caption{The cost ratio increases and then decreases with increasing $k$. Shown for the \texttt{mice-protein} dataset.} 
    \label{fig:cost_ratio_decreases}
    \end{figure}

We had noted in Section \ref{sec:expclust_obs} that cost ratios seem to increase with increasing $k$ - consider the case of the \texttt{mice-protein} dataset in Figure  \ref{fig:expclust}(a). This is counter-intuitive since distances to centroids should decrease due to a finer partitioning of the space, which should drive down $J_{Ex}$ in Equation \ref{eqn:kmeans}. As shown in Figure \ref{fig:cost_ratio_decreases}, this indeed does happen after a temporary rise in cost-ratios; the figure shows the \texttt{mice-protein} dataset again, but now $k$ goes up to $80$ rather than $10$ as in the main paper.

\end{document}